%% file: m1722.tex
\documentclass{ecai} 

\usepackage{amsmath}
\usepackage{algorithm}
\usepackage[noend]{algpseudocode}

\makeatletter
\def\BState{\State\hskip-\ALG@thistlm}
\makeatother

\usepackage{graphicx}
\usepackage{latexsym}
\usepackage[utf8]{inputenc} 
\usepackage[T1]{fontenc}    
\usepackage{hyperref}       
\usepackage{url}            
\usepackage{booktabs}       
\usepackage{amsfonts}       
\usepackage{nicefrac}       
\usepackage{microtype}      
\usepackage{xcolor}         
\usepackage{float} 
\usepackage{multirow}
\usepackage{graphicx}

\begin{document}
\begin{frontmatter}


\paperid{1722} 

\title{Large Language Model Prompting With Episodic Memory}

\author[A]{\fnms{Dai}~\snm{Do}\thanks{Corresponding Author. Email: v.do@deakin.edu.au.}}
\author[B]{\fnms{Quan}~\snm{Tran}}
\author[A]{\fnms{Svetha}~\snm{Venkatesh}}
\author[A]{\fnms{Hung}~\snm{Le}}

\address[A]{Applied AI Institute, Deakin University, Australia}
\address[B]{ServiceNow Research, USA}

\input{paper_abstract}
\end{frontmatter}
\section{Introduction}
\input{paper_intro}
\section{Method}
\input{paper_method}

\section{Experiments}
\input{paper_exp}

\section{Related Works}
\input{paper_related}

\section{Discussion}
\input{paper_discuss}

\bibliographystyle{abbrv}
\bibliography{ref}

\appendix
\input{paper_appendix}

\end{document}

%% file: paper_abstract.tex
\begin{abstract}
   Prompt optimization is essential for enhancing the performance of Large Language Models (LLMs) in a range of Natural Language Processing (NLP) tasks, particularly in scenarios of few-shot learning where training examples are incorporated directly into the prompt. Despite the growing interest in optimizing prompts with few-shot examples, existing methods for prompt optimization are often resource-intensive or perform inadequately. In this work, we propose PrOmpting with Episodic Memory (POEM), a novel prompt optimization technique that is simple, efficient, and demonstrates strong generalization capabilities. We approach prompt optimization as a Reinforcement Learning (RL) challenge, using episodic memory to archive combinations of input data, permutations of few-shot examples, and the rewards observed during training. In the testing phase, we optimize the sequence of examples for each test query by selecting the sequence that yields the highest total rewards from the top-k most similar training examples in the episodic memory. Our results show that POEM outperforms recent techniques like TEMPERA and RLPrompt by over 5.3\% in various text classification tasks. Furthermore, our approach adapts well to broader language understanding tasks, consistently outperforming conventional heuristic methods for ordering examples.

\end{abstract}

%% file: paper_intro.tex

The recent rapid advancements in Language Models (LMs) have underscored the increasing significance of utilizing pre-trained language models, especially when paired with appropriate prompts, as evidenced by seminal works \cite{brown2020language, chowdhery2023palm, fedus2022switch}. As Language Models increase in parameter count, they unveil new capabilities such as In-Context Learning (ICL) \cite{brown2020language}, enabling LLMs to tackle tasks with just a few example demonstrations in the prompt. ICL offers a data-efficient approach for performing NLP tasks, achieving remarkable few-shot performances across many downstream tasks  \cite{li2021prefix, liu2021makes, liu2021p}.


However, the prompt content and ICL examples necessitate meticulous tuning to ensure consistent performance across various tasks. To optimize prompt contents, early attempts focus on tuning the embeddings via gradient descent ("soft prompts", \cite{lester2021power, liu2021p}). Unfortunately, soft prompts require gradients from LLMs to construct prompts and often face challenges with interpretability and quality \cite{khashabi2021prompt, lester2021power}. Additionally, they struggle to handle ICL examples within the prompt, and thus can only be used for zero-shot prompting. Consequently, the current state-of-the-art has shifted towards discrete prompt optimization \cite{deng2022rlprompt, zhang2022tempera}, which enhances interpretability and permits ICL optimization. 

Selecting the right in-context examples and their orders in prompts is crucial for ICL optimization \cite{liu2021makes}. This task is challenging due to the vast array of possible combinations and diverse instructions \cite{lu2021fantastically}. Furthermore, the arrangement of these examples may introduce biases, including majority, recency, and primacy biases \cite{lu2021fantastically, zhao2021calibrate, pezeshkpour2023large}. Although reasonable example selection can be achieved through nearest neighbor retrieval \cite{liu2021makes}, determining the optimal order of examples remains an unresolved research challenge.

Initial efforts employed heuristic rules to rank examples in descending or ascending order based on their similarity to the test instance \cite{liu2021makes, lu2021fantastically}. More recent work attempts to use LLMs as black-box optimizer \cite{sun2022black, prasad2022grips}, calibration \cite{zhao2021calibrate} or applies RL-based method for generating prompts with ICL examples \cite{zhang2022tempera}. 

Discrete prompt optimization presents its own set of challenges. Heuristic methods lack optimization principles, leading to success in some cases but failure in others \cite{liu2021makes}. Black-box methods are guided-optimization and gradient-free. However, they are query-agnostic, thus failing to incorporate any query-related context into the prompt, which can lead to downstream performance degradation. 
Moreover, they often require additional LLM computation for prompt generation, resulting in extensive resource usage \cite{yang2023large, pryzant2023automatic, zhou2022large}. Although using RL-based methods for prompt editing sequentially presents a potential solution that does not require extra LLM for prompt generation \cite{zhang2022tempera}, their slow convergence and intrinsic complexity hinder effectiveness. 

\begin{figure*}[htp]
    \centering
    \includegraphics[width=1\textwidth]{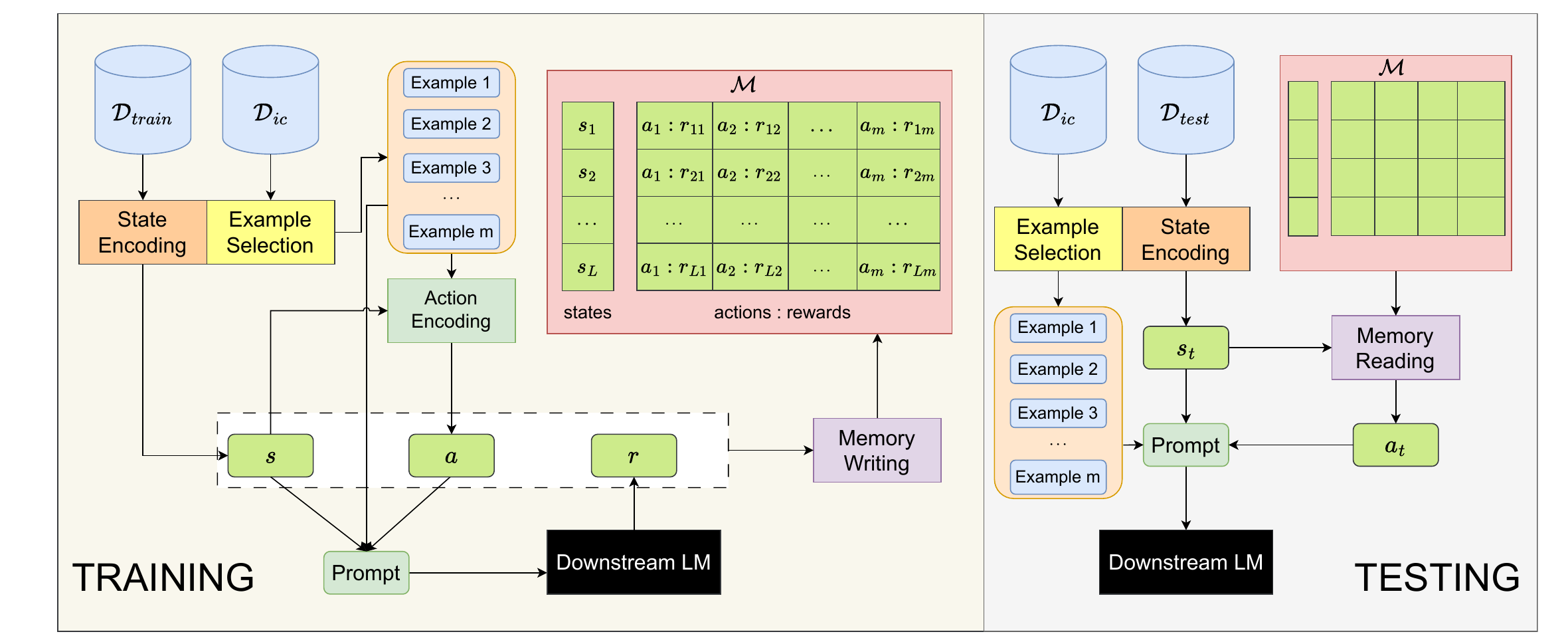}
    \caption{POEM Architecture. Training (left): In this phase, we select examples from the in-context dataset $D_{ic}$. The training query and the ICL example ordering are encoded into $s$ and $a$, respectively, and are used to construct a prompt for each training query. Then, we receive a reward $r$ by feeding the prompt to the downstream language model (LM), and we store the state, action, and reward in memory $\mathcal{M}$ using \textbf{Memory Writing} (Eq.~\ref{Algo1}). Testing (right): During this phase, for each testing query $s_t$, we conduct \textbf{Memory Reading} using nearest neighbor estimation to get the action with the highest estimated value (Eq.~\ref{Algo2}). We then build the prompt for the test query by producing the ICL examples that correspond to the best ordering action $a_t$.
    }
    \label{fig:poem}
    \vspace{10pt}
\end{figure*}

In this paper, we propose a novel and efficient memory-based approach to optimize the order of ICL examples within LLM prompts. Drawing inspiration from the rapid, model-free, and instance-based learning evident in the hippocampus region of the human brain \cite{lengyel2007hippocampal}, our method eliminates the necessity for complex reinforcement learning optimization while being more reliable than heuristic methods through performance-driven optimization. Leveraging episodic control mechanisms in reinforcement learning \cite{blundell2016model, le2022neurocoder, le2022episodic}, we formulate each evaluation of training data as an episode and utilize an episodic memory to store the performance of any combination of the training data and ICL orderings. By sampling and evaluating certain training inputs and ICL order pairs, we avoid exhaustive searches across all data-ICL order combinations, which is particularly beneficial when LLM evaluation is costly. During testing, this memory serves as a non-parametric nearest-neighbors model, utilizing the recorded performance of similar training data to determine the optimal order for the testing data.

To ensure the robust generalization of our episodic memory, we devise specialized representations for both the text input and the ordering of ICL examples. Specifically, we encode the input using the last hidden states of a pre-trained language model, ensuring high-quality similarity-based retrieval during testing. 
Moreover, directly encoding the permutation as a sequence of ICL examples would result in a vast search space. Suppose there are $M$ training samples in total and we can select $m$ examples for each prompt, we would have $\frac{M!}{(M-m)!}$ possible ICL arrangements. Instead, we represent the ordering as a permutation of the similarity rank of the in-context examples. Specifically, given a specific ordering of ICL examples, we (1) gauge the similarity between the examples to the testing input, (2) assign a rank to each example based on the similarity, and (3) encode the ICL ordering as the sequence of the ranks. Our encoding operates within the similarity rank space rather than the text space, effectively reducing the search space from $\frac{M!}{(M-m)!}$ to $m!$. More importantly, this approach also encourages generalization as the permutation focuses solely on the arrangement of the example rank rather than specific content.


In summary, we propose POEM, a method that optimizes in-context example ordering during test time using an Episodic Memory. It utilizes a similarity ranking to encode example orders based on their proximity to the test instance. 
In our few-shot classification experiments across seven datasets, POEM outperforms TEMPERA on six of these datasets. Additionally, POEM achieves an average performance improvement of 13.4\% over RLPrompt, demonstrating a significant advantage. For tasks requiring larger LLMs, such as Commonsense Reasoning and Question Answering, POEM consistently outperforms heuristic baselines across all four LLMs tested.

%% file: paper_method.tex
\subsection{Problem Formulation}

\textbf{Few-shot text classification.} We follow the standard few-shot setting for downstream tasks of language models \cite{brown2020language}. In this setup, a pretrained language model $\mathcal{L}$ is paired with a task-specific dataset $\mathcal{D}$, where $\mathcal{Y}$ represents the label space. $\mathcal{Y}$ may vary in form; it can be categorical, as in classification tasks, or sequential, as in question answering and commonsense reasoning tasks.  For classification tasks, we randomly assemble a dataset of $L = g * \mathcal{G}$ samples, with $g$ as the samples per label and $\mathcal{G}$ as the total labels. In cases without categorical labels, we randomly sample $L$ instances from $\mathcal{D}$. This forms the training dataset, denoted as $D_{train} = \{x_i,y_i\}_{i=1}^{L}$, while a separate hold-out set $D_{test}$ is reserved for evaluation. 

\textbf{In-context Learning.} Following GPT-3 paper \cite{brown2020language}, \textit{In-context Learning is a paradigm that allows language models to learn tasks given only a few examples in the form of demonstration.} Given a test sentence $x_{test}$, template-based construction $\Psi$, along with an optional task description $t_{desc}$ and a set of in-context examples $\mathcal{T} = \{x_i, y_i\}_{i=1}^{m}$, denoting $\Gamma$ as the prompt construction function, we can formulate an input prompt $p$ as follows:

\begin{equation}
\begin{aligned}
p &= \Gamma \left(t_{desc}, \mathcal{T}, x_{test} \right) \\
&= t_{desc} \oplus \Psi\left(x_1,y_1\right) \oplus \dots \oplus \Psi\left(x_{m},y_{m}\right) \oplus \Psi\left(x_{test},*\right) 
\end{aligned}
\label{eq1}
\end{equation}
where $\oplus$ is the concatenate function and $m$ is the number of in-context examples for each prompt. In line with the few-shot setting \cite{zhang2022tempera,deng2022rlprompt}, we consider an in-context set of $M$ samples, denoted as $\mathcal{D}_{ic} \in \mathcal{D}$, excluding the few-shot training data. The in-context examples will be sampled only from this set.

In-context learning facilitates the construction of the task's output distribution $p_{LM}\left(y|x,p\right)$, where $x$ represents the input string and $y$ represents the output string. This powerful approach allows in-context learning to play an active role in shaping the selection and arrangement of the demonstration set $\mathcal{T}$ within the input prompt $p$. By carefully curating and organizing these demonstrations, in-context learning can significantly enhance the model's ability to perform effective optimization, ultimately leading to more accurate and reliable task outcomes.

\textbf{Reinforcement Learning Formulation.} We formulate in-context prompt optimization as an RL problem where a state $\textbf{s}=\mathcal{E}\left(x\right)$ is the embedding of the input $x$, where $\mathcal{E}$ is a pre-trained encoder. During training, given a set of $m$ in-context examples, the RL agent selects one of the possible permutations as its action $a$ from the action space $\mathcal{A}$. We then construct the prompt $p$ using the default task description/instruction (if any), combined with the $a$-ordered in-context examples, and query it to the downstream LM to get the reward $r$. The goal of the agent is to learn a policy that maximizes the episodic return $R_t = \sum_{h=0}^{H-t}  r_{t+h}$, where $H$ is the time step at which the episode ends. To simplify the formulation, our episode consists of only one step, during which the sole action is selecting the permutation of the in-context examples.

\subsection{Prompting with Episodic Memory} \label{promptingwithepisodicmemory}

\begin{algorithm}[t]
\caption{In-context examples order optimization with Episodic Memory}\label{euclid}
\begin{algorithmic}[1]
\Require{Language model $\mathcal{L}$, State encoder $\mathcal{E}$, Training set $\mathcal{D}_{train}$, Evaluation set $\mathcal{D}_{test}$, In-context set $\mathcal{D}_{ic}$, Number of iterations $N$, Episodic Memory $\mathcal{M}$, Number of neighbors $k$, Task description $t_{desc}$, Template transformation $\Psi$}, Number of in-context examples per prompt $m$, Prompt construction function $\Gamma$
\State Initialize $\mathcal{M}=\emptyset$

\State \textbf{Training:}
\For {episode $n=1$ to $N$ \do}
\State Random sample batch $\mathcal{B}$ $\sim$ $D_{train}$
\For {x $\in \mathcal{B}$ \do}
\State Receive state $s = \mathcal{E} \left( x \right)$
\State Receive in-context examples $\mathcal{T}_s =\Omega \left( s,\mathcal{D}_{ic} \right)$
\State Select permutation $a$ $\leftarrow$ linearly decaying $\epsilon$-greedy policy 
\State(Eq. \ref{epsilongreedy}) using $\mathcal{M}$ (Eq.~\ref{Algo2})
\State Reorder the in-context examples set $\mathcal{T}_s^a = \nabla\left(\mathcal{T}_s, a\right)$
\State Get prompt $p=\Gamma\left(t_{desc}, \Psi \left(\mathcal{T}_s^a\right)\right)$
\State Receive reward $r$
\State Update $\mathcal{M}$ using \textbf{Memory writing} with $r$ (Eq. ~\ref{Algo1})
\EndFor
\State \textbf{endfor}
\EndFor
\State \textbf{endfor}

\State$\textbf{Testing:}$
\For {$x_t \in \mathcal{D}_{test}$ \do}
\State Receive state $s_t = \mathcal{E}\left(x_t\right)$
\State Receive in-context example set $\mathcal{T}_{s_t} = \Omega \left( s_t,\mathcal{D}_{ic} \right)$
\State Obtain $\widehat{\mathcal{M}}\left(s_t, a\right)$ with \textbf{Memory reading} (Eq. ~\ref{Algo2})
\State Obtain $a_t$ via Eq. ~\ref{obtaina}
\State Reorder the in-context examples set $\mathcal{T}_{s_t}^{a_t} = \nabla\left(\mathcal{T}_{s_t},a_t\right)$
\State Get prompt $p=\Gamma\left(t_{desc}, \Psi \left(\mathcal{T}_{s_t}^{a_t}\right)\right)$  
\State Get prediction $\hat{y} = \mathcal{L}\left(x_t, p\right)$
\EndFor
\State \textbf{endfor}

\end{algorithmic}
\label{Overallalgo}
\end{algorithm}

In this section, we present the architecture of our episodic memory. The memory is structured as a dictionary, storing the embeddings of training sentences (states) as keys. Each key's value is another dictionary, mapping a permutation (action) to its respective reward. We denote a key by $s_i$, and for each key, $a_j$ represents the  $j$-th permutation, while $r_{ij}$ denotes the associated reward. Our episodic memory $\mathcal{M}$ can be represented by the following structure:

\begin{equation}
\mathcal{M}=\{s_i: \{a_1:r_{i1}, a_2:r_{i2}, \dots, a_p:r_{ip}\}\}_{i=1}^L
\end{equation}
Here, $p$ signifies the total number of permutations available for the in-context examples (with $m$ examples per prompt, $p=m!$). $L$ is the total of the stored states in the episodic memory. An illustration of our memory architecture is given in Fig. \ref{fig:poem}.  Below are the detailed components of the memory.

\textbf{State representation} Obtaining accurate and meaningful text representations for the state is crucial for both memory storage and efficient retrieval. In our approach, we leverage the power of the encoder $\mathcal{E}$, specifically utilizing the SentenceTransformers model \cite{reimers2019sentence}. This encoder is designed to generate high-quality sentence embeddings that can be effectively applied across a wide range of NLP tasks, ensuring that the textual data is both rich in information and can be used effectively in various NLP tasks.

\begin{table*}[htp]
    \centering 
    \begin{tabular}{p{2.5cm}p{1.1cm}p{1.1cm}p{1.1cm}p{1.2cm}p{1.1cm}p{1.1cm}p{1.1cm}p{1.1cm}} 
    \toprule
     & SST-2 & Boolq & CR & AG News & IMDB & QNLI & COLA & \textbf{Average}\\
    \midrule
    Finetune & 80.6(3.9) & 55.3(3.1) & 73.3(7.5) & \textbf{84.9(3.6)} & \underline{85.4(5.2)} & \underline{55.4(1.6)} & 55.6(9.9) & 70.1\\
    \midrule
     Manual Prompt & 82.8 & 61.0 & 79.6 & 76.9 & 85.0 & 51.0 & 32.0 & 66.9\\
     In-Context Demo. & 85.8(0.7) & 58.3(0.4) & 85.5(1.5) & 74.9(0.8) & 69.8(0.6) & 53.5(0.5) & 56.0(1.2) &69.1\\
     Instruction & 89.0 & 61.0 & 80.8 & 54.8 & 89.0 & \textbf{56.0} & \underline{61.0} & 70.2\\
     Black-box Tuning & 89.1(0.9) & 53.3(1.7) & 87.4(1.0) & \underline{83.5(0.9)} & 68.9(5.1) & 50.8(0.6) & 51.4(2.1) & 69.2\\
     RLPrompt & 87.7(3.6) & 41.6(3.0) & 90.4(1.7) & 73.8(5.8) & 57.3(8.9) & 50.1(0.8) & 51.6(1.5) & 64.6\\
     TEMPERA & \underline{90.5(1.6)} & 54.9(5.8) & \underline{91.1(1.1)} & 81.6(2.5) & 85.3(2.8) & 51.6(1.0) & 54.2(5.5) & \underline{72.7}\\
    Random Ordering & 81.5(0.9) & \underline{62.7(0.9)} & 87.2(0.4) & 57(0.2) & 74.1(0.5) & 53.9(0.4) & 41.63(1.3) & 65.4\\
    \midrule
    POEM (Ours) & \textbf{93.4(0.2)} & \textbf{66.1(0.1)} & \textbf{92.6(0.2)} & 80.3(1.1) & \textbf{90.9(0.1)} & 54.3(0.2) & \textbf{68.4(0.4)} & \textbf{78.0}\\
    \bottomrule
    \end{tabular}
    \caption{Accuracy and standard derivation (if available) of different baselines on few-shot classification over 4 seeds. Some methods like Manual Prompt produce the same results across seeds, and thus, have no standard derivation to report. The highest accuracy is \textbf{bolded} and the second-highest accuracy is \underline{underlined}. The last column shows the average accuracy across all datasets in this table.}
    \label{class_results}
    \vspace{10pt}
\end{table*}

\textbf{Example selection} 
To simplify the optimization, we do not aim to optimize the example selection process. Therefore, following prior work \cite{liu2021makes},
from the in-context dataset $D_{ic}$, given input $x$, we select $m$ in-context examples that are semantically closest to $x$. We measure the semantic similarity between in-context example $x^i_{ic}$  and the input query $x$ using Cosine Similarity:

\begin{equation}
\small
    CS \left(x, x_{ic}^i\right) = \frac{s \cdot s_{ic}^i}{||s|| ||s_{ic}^i||} 
\label{eq2}
\end{equation}

In the context of a \textit{few-shot classification} task, where label biases can significantly influence the prompting outcome, it is crucial to maintain an equal number of examples for each label. To address this, we propose the following strategy: if there are $\mathcal{G}$ unique labels in $\mathcal{D}$, and $\mathcal{G} < m$, we select $\left\lfloor \frac{m}{\mathcal{G}} \right\rfloor$ closest samples from each label. Conversely, if $\mathcal{G} \geq m$, we iteratively choose one sample for each label until we have sufficient in-context examples.
We denote the above process as the function $\Omega$ that retrieves context examples for each state $s$:

\begin{equation}
    \mathcal{T}_s = \Omega \left(s, \mathcal{D}_{ic}\right)
\label{prepare}
\end{equation}
where $\mathcal{T}_s$ is the set of in-context example for the state $s$

\begin{figure}[t]
    \centering
    \includegraphics[width=\linewidth]{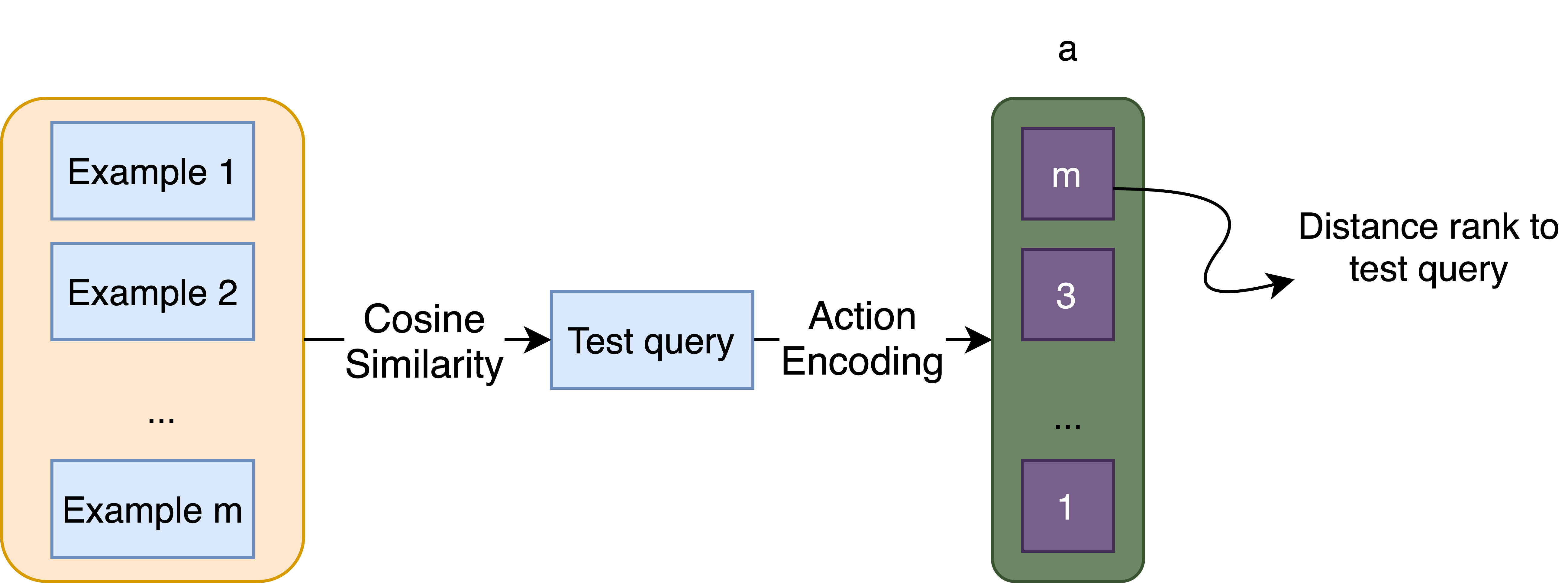}
    \caption{Illustration of an action being encoded.}
    \label{fig:encodeaction}
    \vspace{20pt}
\end{figure}

\textbf{Action encoding} An action refers to a specific arrangement of in-context examples within $\mathcal{T}_s$.  As mentioned in the introduction, a naive approach of encoding the action as a sequence of ICL examples will lead to a huge action space. Therefore, we propose to encode the action as a sequence of similarity ranks. Concretely, for each state $s$, we measure the semantic similarity between it and the states of the in-context examples using Eq. \ref{eq2}. Next, we rank each example according to its similarity to the input query and encode the ICL ordering as a sequence of these ranks. 

For illustration, consider the action encoding process illustrated in Fig. \ref{fig:encodeaction}. First, we compute the cosine similarity between each in-context example and the test query, which ranks these examples based on their distance. To encode an action, we create a permutation of these rankings, resulting in a unique action sequence. For instance, in Fig. \ref{fig:encodeaction}, the action $a = \left(m, 3, \dots, 1\right)$ represents a specific permutation of the in-context examples. Here, the first example is the farthest from the test query, the second is the third closest, and so on, with the $m$-th position being the closest to the test query. Each action is thus a vector with $m$ elements. This encoding captures the relational similarity among the in-context examples and the test query, with the action space size being $m!$.

Given a set of in-context examples  $\mathcal{T}_s$ and an action $a$, we define a permutation function $\nabla$, used to reorder the elements of $\mathcal{T}_s$ according to the sequence specified by $a$. The action $a$ represents a specific ordering of indices that correspond to the elements in $\mathcal{T}_s$. Formally, we have:

\begin{equation}
    \mathcal{T}^a_s = \nabla \left( \mathcal{T}_s, a\right)
\end{equation}
where $\mathcal{T}^a_s$ represents the sequence of in-context examples rearranged in accordance with action $a$.

\textbf{Reward design} For \textit{classification} tasks that use Masked LM such as RoBERTa \cite{liu2019roberta}, for each query $x$, we define the reward based on the log probability of the output label of the model $\log P_{\mathcal{L}} \left(\hat{y} | x, p\right)$

\begin{equation}
    \footnotesize
    r \left( c, x, p \right) = \lambda_1 \log \mathcal{P}_{\mathcal{L}} \left( \hat{\mathrm{y}}_c \mid x, p \right) - \lambda_2 \max_{c \neq c'} \log \mathcal{P}_{\mathcal{L}} \left( \hat{\mathrm{y}}_{c'} \mid x, p \right)
\label{eq3}
\end{equation}
where $\lambda_1 > 0$ and $\lambda_2 > 0$ are the two balance hyperparameters for the positive and negative terms respectively.

For \textit{generative} tasks (e.g: Commonsense Reasoning, Question Answering) that use Causal LM such as Llama \cite{touvron2023llama}, where ground truth label $c$ is a sequence of tokens, we define the reward as:

\begin{equation}
\footnotesize
\begin{split}
    r \left( c, x, p \right) &= \lambda_1 \sum_{i=1}^{|c|} \log \mathcal{P}_{\mathcal{L}} \left( i \mid x, p \right) - \\
    &\qquad \lambda_2 \max_{c \neq c'} \sum_{j=1}^{|c'|} \log \mathcal{P}_{\mathcal{L}} \left( j \mid x, p \right)
\end{split}    
\label{sequencereward}
\end{equation}
In the equations above, $c'$ is the token/sequence that has the highest probability. In Eq. \ref{sequencereward}, $i$ and $j$ are the tokens of two sequences $c$ and $c'$, respectively. Intuitively, for classification tasks, the reward is positive when the prediction is correct and negative otherwise.

For tasks that use Causal LM and Exact Match as evaluation metrics, we define the reward as:

\begin{equation}
    r \left( c, x, p \right) =\left\{
  \begin{array}{@{}ll@{}}
    1 , & \text{if}\ c=c' \\
    0, & \text{otherwise}
  \end{array}\right.
\end{equation}

\subsection{Episodic Memory Operations}

Our memory operation involves two main phases. \textbf{Training}: The RL agent interacts with the LLM to try actions and collect rewards for $\left(s,a\right)$ pairs to fill the memory. \textbf{Testing}: The RL agent uses the memory to determine the ICL ordering for testing data.

\textbf{Training} In this phase, we collect and store the rewards defined above for state-action pairs \(\left(s,a\right)\). Given an input query \(x \in \mathcal{D}_{\text{train}}\), we obtain its representation using the pre-trained encoder, \(s = \mathcal{E}\left(x\right)\). The action at training time is selected via a linearly decaying \(\epsilon\)-greedy policy, defined as:

\begin{equation} \label{epsilongreedy}
    \epsilon_t = \epsilon_{\text{initial}} - \left(\epsilon_{\text{initial}} - \epsilon_{\text{final}}\right) \cdot \frac{t}{N}    
\end{equation}

where \(\epsilon_t\) is the value of \(\epsilon\) at episode \(t\), N is the total number of training iterations, \(\epsilon_{\text{initial}}\) and \(\epsilon_{\text{final}}\) are the initial and final values of \(\epsilon\), respectively.

Given the a pair of state and action $(s, a)$ and the reward $r$, following \cite{blundell2016model}, our Episodic Memory $\mathcal{M}$ is updated using \textbf{Memory writing} as follows:

\begin{equation}
  \mathcal{M} \left( s, a\right) \leftarrow \left\{
  \begin{array}{@{}ll@{}}
    r , & \text{if}\ \left( s, a \right) \notin \mathcal{M} \\
    \max{\big\{ \mathcal{M} \left( s, a \right), r \big\}}, & \text{otherwise}
  \end{array}\right.
  \label{Algo1}
\end{equation}

It is noted that the stored rewards never decrease, indicating a focus on high-return actions. In our setting, we use the highest possible reward that the downstream LM can achieve for the pair $(s,a)$ to estimate the value of using ICL ordering $a$ for input $x$.   The motivation behind this is to emulate the brain's specialized learning mechanisms that exploit predictable patterns in the environment, enabling rapid learning from high-return actions recorded in memory. 

We limit the size of our memory to be equal to the size of the training dataset $\mathcal{D}_{train}$. In the few-shot scenario where there are limited training samples, it is guaranteed that our memory will not be overflowed. However, in other settings where $\mathcal{D}_{train}$ is too big, our memory can have a fixed smaller size, and when a new state-value pair has to be introduced, the least recently used state will be discarded.

\textbf{Testing} In this phase, we select the optimal permutation for each testing query. In reality, it is common to receive novel states (states that are not seen in training). To handle this, we employ a nearest-neighbor estimator. In particular, we obtain the approximated value of a testing state $s_t$ and a candidate action $a$ using the following \textbf{Memory Reading} process \cite{le2021model}:

\begin{equation}
  \widehat{\mathcal{M}} \left( s_t, a\right)=\left\{
  \begin{array}{@{}ll@{}}
    \sum_{i=1}^{k} {\mathcal{M} \left( s^i, a\right)}\frac{CS\left(x_t, x^i\right)}{ \sum_{j=1}^{k} CS\left(x_t, x^j\right)} , & \text{if}\ \left( s_t, a \right) \notin \mathcal{M} \\
    \mathcal{M} \left( s_t, a\right), & \text{otherwise}
  \end{array}\right.
  \label{Algo2}
\end{equation}
where $s^i,\  i = 1, \dots , k$ are the $k$ states with the highest similarity to the testing state $s_t$, $CS\left(x_t, x^i\right)$ is the Cosine Similarity between the neighbor $x^i$ to the test query $x_t$. The motivation behind the weighted sum is that we believe the semantic similarity between training data and the test query should correspondingly affect its weight in the ordering process. After having calculated $\widehat{\mathcal{M}}\left(s_t,a\right)$, we get the action $a_t$ for $s_t$ as follow:

\begin{equation} 
    a_t = \arg \max_a \widehat{\mathcal{M}}\left(s_t, a\right)
    \label{obtaina}
\end{equation}


\begin{table*}[t]
\centering

\begin{tabular}{p{1.8cm}p{1.8cm}p{1.3cm}p{1.3cm}p{1.3cm}p{1.3cm}}
\toprule
Model name & Baseline & Hellaswag & PIQA & TruthfulQA & TriviaQA \\
\midrule
& & acc. &acc. &acc. &EM\\
\midrule
& 0-shot& 57.79& 76.38& 25.99& 18.90\\
& Random & 58.59& 77.37& 32.38& 58.65\\
& Ascending& 58.36& 76.22& 26.94 & 57.06\\
& Descending& \underline{59.61}& \underline{78.73}& \underline{38.90}& \underline{59.01}\\
Llama 2 7B & POEM (Ours)& \textbf{59.65}& \textbf{79.27}& \textbf{38.91} &\textbf{59.06}\\
\midrule
 & 0-shot & 60.73 & 77.58&24.63& 27.26 \\
 & Random&  62.40& \underline{79.05}& 32.79&65.98\\
 & Ascending& 61.39&76.77& 26.94& 64.00\\
 & Descending& \textbf{63.17}& 79.00& \underline{40.41}& \textbf{67.13}\\
 Llama 2 13B & POEM (Ours) & \underline{63.13}& \textbf{79.43}& \textbf{40.95}&\underline{67.06}\\
 \midrule
 &0-shot &63.76& 79.27& 28.16&42.69\\
 &Random & 65.59& 80.72 & 38.64&73.09\\
 &Ascending &65.59 &80.79 & 32.24& 72.07\\
 &Descending& \underline{66.99}&\textbf{81.66}& \underline{45.17}& \underline{73.53}\\
 Llama 2 70B&POEM(Ours)&\textbf{67.15}& \underline{81.12}& \textbf{45.71} & \textbf{73.56} \\
 \midrule
 & 0-shot& 65.98& 80.20& 46.26& 32.41\\
 & Random& 66.29& 81.77& 54.01& 63.07\\
 & Ascending& 65.47& 81.50& 50.88& 62.04\\
 & Descending& \underline{67.00}& \textbf{83.51}& \textbf{59.46}& \underline{63.75}  \\
 Mistral 7B& POEM (k=10)&\textbf{67.14} & \underline{83.35}& \textbf{59.46}& \textbf{63.83}  \\
\bottomrule
\end{tabular}
\caption{Comparison between POEM and other heuristic baselines. All baselines (except for 0-shot) use 4 in-context examples for each prompt. \textbf{Bolded} are the best results and \underline{underlined} are the second best. The metric for evaluation for each dataset is written below the dataset names. \textit{acc.} stands for accuracy and \textit{EM} stands for exact match.}
\label{llmtasks}
\vspace{10pt}
\end{table*}

We note that with enough training steps, the actions for each state in the memory are filled to assure Eq.~\ref{obtaina} is valid. In addition, the nearest neighbor estimation here is a different process from the nearest neighbor retrieval in the example selection described in Section 2.2. Algo. \ref{Overallalgo} summarizes the procedure of our method.

%% file: paper_exp.tex
\subsection{Few-Shot Text Classification}

\textbf{Datasets.} For classification tasks, we conduct experiments for sentiment analysis, Natural Language Understanding (NLU), topic classification, and natural language inference datasets. For sentiment analysis, we choose SST-2 \cite{socher2013recursive}, IMDB \cite{maas-etal-2011-learning} and CR \cite{10.1145/1014052.1014073} as datasets. For NLU task, we select COLA \cite{wang2018glue}. For Reading Comprehension, we choose Boolq \cite{clark2019boolq}. For topic classification, we use AG News \cite{zhang2015character}. For Natural Language Inference (NLI), we choose QNLI \cite{wang2019superglue}. The statistics, manual templates, and label words of this dataset are shown in Appendix A.3. 

\textbf{Baselines.} For few-shot classification tasks, we compare our work with previous continuous and discrete prompt optimization methods. \textit{Finetuning} is the method that finetunes the entire language model with a classification head using the few-shot dataset. For \textit{Manual Prompt}, we use the hand-written prompts from \cite{schick2020exploiting}. \textit{In-context Demonstration} randomly selects one example and concatenates it with the test query. \textit{Black-box Tuning} \cite{sun2022black} combines a discrete component optimized through non-gradient methods and a soft component refined via gradient descent. \textit{RLPrompt} \cite{deng2022rlprompt} generates discrete prompt tokens using RL framework. \textit{TEMPERA} \cite{zhang2022tempera} trains a RL agent to edit prompt sequentially. To ensure a fair comparison, we rerun these methods under our setting. Besides complicated baselines, we also investigate a simple heuristic baseline. \textit{Random Ordering} adopts nearest neighbor example selection, randomly permutes the in-context examples, and concatenates with the test query.


\textbf{Experiment Setup.} We use RoBERTa-large \cite{liu2019roberta} as the downstream LM. To ensure a fair comparison, we follow the same setting from \cite{deng2022rlprompt, zhang2022tempera} by testing all baselines on a few-shot text classification setting. Training dataset $\mathcal{D}_{train}$  has 16 examples per class, and we sample another 16 data points as in-context dataset $\mathcal{D}_{ic}$ ($M=16$). For reporting the testing results, we make use of the models having the highest performance on the validation set and do inference on the test set $\mathcal{D}_{test}$ provided by the task. We use $m=4$ examples for each prompt as in prior works \cite{zhang2022tempera}. For each run, we use the same training and in-context samples for all baselines. At test time, we select the number of nearest-neighbor $k=10$ for POEM.

\textbf{Results.} We present our few-shot text classification results over 4 runs in Tab. ~\ref{class_results}. We can see that on most tasks, POEM outperforms previous baselines by a large margin. For example, we have a 2.9 \% absolute gain on SST-2 task (over TEMPERA), and 5.6 \% on IMDB, and the performance is almost comparable to finetuning the LM on QNLI task. Especially, for NLI and NLU tasks, on harder datasets like Boolq and COLA, we have a significance gain of 11.2 \% and 14.2 \% respectively. We also see that POEM has a much smaller variance between different runs than all other baselines, which indicates that it is more stable across different datasets. Unlike TEMPERA, which requires sequential modification of in-context examples, our method, POEM, enables one-step optimization. Our approach not only simplifies the process but also overcomes TEMPERA's slow inference involving several numbers of edits for each prompt. POEM also outperforms Random Ordering significantly, by nearly 13\% on average, highlighting the importance of ICL optimization. More comparisons with advanced heuristic approaches are given in Appendix A.4.

\subsection{General Language Understanding Tasks}

We further extend experiments of our method to general language understanding tasks that require stronger downstream LLM to generate the answers.

\textbf{Datasets.} We measure performance on two main tasks with four datasets, categorized as follows: \textbf{Commonsense Reasoning.} Hellaswag \cite{zellers2019hellaswag}, PIQA \cite{bisk2020piqa}; \textbf{Question Answering.} TruthfulQA-mc1 \cite{lin2021truthfulqa}, TriviaQA \cite{joshi2017triviaqa}. For more complex tasks such as Question Answering and Reading Comprehension, which involve multiple text fields, we employ a reranking approach based on the field containing the most relevant information. In cases like TruthfulQA, where certain fields like \textit{type} and \textit{category} lack semantic significance, we measure textual distances between examples using the \textit{question} field, which typically encapsulates the query to be answered. For more detailed information on datasets and the fields used for retrieval, please refer to Appendices A.3 and A.4.

\textbf{Baselines.} We note that complicated optimization baselines such as RLPrompt and TEMPERA have not been designed and applied to this task. Therefore, we compare POEM with several heuristic baselines. \textit{0-shot} only includes the test query. \textit{Random Ordering} randomly permute the examples. \textit{Ascending Ordering}  arranges in-context examples by increasing relevance, placing the most pertinent example closest to the test instance. In contrast, \textit{Descending Ordering} positions examples from most to least relevant in relation to the test instance. 

\textbf{Experiment Setup.} We use Llama-2-7b-chat, Llama-2-13b-chat, Llama-2-70b-chat \cite{touvron2023llama} with default parameters. For all datasets, similar to classification tasks, we randomly sample 16 examples for training and another 16 examples to form an in-context dataset. For all baselines (except for 0-shot), we use $m=4$ in-context examples for each test query. At test time, we select the number of nearest neighbors $k=10$. Baselines that use ICL examples share the same example selection mechanism described in the Method section. The details of parameters used for these language models can be seen in Appendix A.2. 

\textbf{Results.} We present our results for general language understanding tasks in Tab. ~\ref{llmtasks}. Overall, it is clear that POEM shows superior results compared to Random and 0-shot baselines. This suggests that our approach is beneficial for optimizing ICL example ordering. Compared to heuristic methods, in the context of 4 examples and for these datasets, POEM performs slightly better. This is because in these cases, the LLMs seem to favor the examples to be ordered descendingly. However, we note that POEM performs better in 9 out of 12 different settings compared to the Descending baseline. We also would like to highlight that POEM consistently shows up in the top 2 best baselines across all datasets. We acknowledge the challenges in replicating Llama's evaluation setup due to the unavailability of their system prompts. Consequently, to maintain simplicity and consistency, our study does not incorporate specific instructions into the prompts for LLMs. Instead, we apply uniform templates across all baselines. This approach ensures fair comparison between POEM and other baselines, thus showing the impact of in-context example reordering. For further details on the templates for this task, please see Appendix A.4.

\subsection{Ablation Studies}
\input{ablation}

%% file: ablation.tex
\subsubsection{Analysis of Efficiency}

\begin{table}[t]
\centering
\begin{tabular}{{p{2.0cm}p{0.8cm}p{0.8cm}p{0.8cm}p{1.1cm}}}
\hline
 &  SST-2& CR & COLA & AG News\\
\hline
Imbalanced labels & \textbf{93.5} &92.2 &58.6 & 69.7\\
Balanced labels& 93.4&\textbf{92.6} &\textbf{68.4} & \textbf{80.3} \\
\hline
\end{tabular}
\caption{Average accuracy over 4 runs on imbalanced in-context examples selection.}
\label{imbalancedlabels}
\end{table}

\begin{table}[t]
\centering
\begin{tabular}{{p{1.5cm}p{1.0cm}}}
\hline
 Algorithms & Accuracy \\
\hline
 RLPrompt & 52.6 \\
 TEMPERA & 88.0 \\
 POEM & \textbf{93.4}\\
 \hline
\end{tabular}
\caption{Accuracy (60 iterations) of POEM, RLPrompt and TEMPERA on SST-2 dataset.}
\vspace{10pt}
\label{afterfewiter}
\end{table}

\begin{table}[t]
\centering
\begin{tabular}{{p{2cm}p{1.1cm}p{1.1cm}p{1.1cm}p{1.1cm}}}
\hline
 &  SST-2& CR & COLA & AG News\\
\hline
Naive Action & 91.0 &91.4 &68.0 & 79.6\\
Rank Action & \textbf{93.4}&\textbf{92.6} &\textbf{68.4} & \textbf{80.3} \\
\hline
\end{tabular}
\caption{Ablation on action encoding. Average accuracy over 4 runs.}
\label{ablation1}
\end{table}

\begin{table}[t]
\centering
\begin{tabular}{{p{1.5cm}p{2.5cm}p{1.0cm}}}
\hline
 Algorithms & Training time (min) & Accuracy \\
\hline
 RLPrompt & 3100 & 87.7  \\
 TEMPERA & 3208 & 90.5 \\
 POEM & \textbf{21} & \textbf{93.4}\\
 \hline
\end{tabular}
\caption{Comparison between POEM, RLPrompt and TEMPERA in terms of training time until convergence (minutes, the smaller the better) and accuracy of SST-2 dataset.}
\vspace{10pt}
\label{analysis1}
\end{table}

We provide empirical evidence for our claim of POEM being fast and efficient. We compare performance and runtime on SST-2 dataset \cite{socher2013recursive} between POEM, RLPrompt and TEMPERA. For a fair evaluation, all experiments were conducted with one identical GPU Nvidia A100. As shown in Tab. \ref{analysis1}, the training time (until convergence) of POEM is approximately 150 times faster than that of TEMPERA and RLPrompt while achieving better accuracy. We also compare the performance of POEM with that of TEMPERA and RLPrompt after 60 iterations. The results in Tab. \ref{afterfewiter} reveal that POEM has attained a state of convergence, demonstrating the effectiveness of its learning algorithm within the given iteration frame. On the other hand, TEMPERA exhibits ongoing optimization efforts beyond the 60-iteration mark. This observation leads us to posit that TEMPERA's decent performance is attributable to its initial prompt construction methodology. RLPrompt, however, is still in the early stages of training, reflected by its near-random accuracy. It is expected that RLPrompt will need more iterations to properly improve its prompt generation process.

\subsubsection{Analysis of POEM's Components}
\textbf{Action encoding.} We aim to investigate further how our action encoding contributes to our framework. To achieve this, we design a naive action encoding for a sequence of \( m \) examples. This approach results in an action space of \(\frac{M!}{(M-m)!}\) rather than \( m! \), where \( M \) denotes the total number of in-context examples and \( m \) represents the number of examples per prompt. As demonstrated in Tab. \ref{ablation1}, the absence of our similarity-ranked encoding leads to a noticeable decrease in performance across all classification tasks. Notably, on the SST-2 dataset, the accuracy drops by 2.4 percentage points, from 93.4\% to 91.0\%. This drop highlights the significant impact of our similarity-ranked encoding on maintaining high performance levels.

\textbf{Imbalanced labels.} For \textit{few-shot classification} task, we aim to investigate how imbalanced in-context example labels can affect performance. Instead of employing our example selection strategy described in Section \ref{promptingwithepisodicmemory} to ensure an equal number of labels within a prompt, we now select the top-$m$ closest examples to the test query regardless of their labels and use them to construct prompts. The results can be seen in Tab. ~\ref{imbalancedlabels}. We can clearly see that for sentiment classification datasets (SST-2 and CR), POEM with imbalanced labels can still perform well. For datasets that require natural language understanding (NLU) like COLA, or those with more labels like AG News, the performances surprisingly dropped significantly. This perhaps indicates that there might be a strong bias coming from the imbalanced example selection for test queries.

\subsubsection{Hyperparameter Sensitivity}

\textbf{Number of iterations.} We present results for different numbers of training iterations, specifically 10, 60 (the default setting), and 120. As shown in Tab. \ref{tableparams2}, increasing the number of iterations tends to improve \textbf{Memory Writing} by allowing POEM to discover more effective permutations. This indicates that more iterations can enhance the model's ability to refine its training. However, we have selected 60 iterations as the default to balance training time with convergence stability, ensuring that the model achieves robust performance without excessive training duration.


\textbf{Size of in-context dataset $\mathcal{D}_{ic}$.} We aim to further investigate how the size of the in-context dataset impacts performance. Intuitively, a larger selection of examples should enhance the effectiveness of the prompts, providing a richer context for generating responses. As shown in Tab. \ref{ablationsize}, increasing the size of the in-context dataset, denoted as $M$, results in a noticeable improvement in performance. This indicates that expanding the dataset contributes positively to the quality of the prompts,

\textbf{Number of nearest neighbors $k$ in memory reading.} We provide results with different number of neighbors to study the robustness of POEM. With $k=6$, $k=10$ and $k=12$, we obtain the results shown in Tab. \ref{tableparams1} for \textit{few-shot classification} datasets, and Tab. \ref{tableparams1_glu} for \textit{general language understanding} datasets. It can be seen that POEM's performance holds for different $k$, which indices our method is relatively stable. We believe the reason behind this stability comes from the weighted sum formula in ~\ref{Algo2}, meaning that the order is predominantly determined by test query's closest neighbors.


\textbf{Number of examples per prompt.} We ablate on the number of examples $m$ used for demonstration in our algorithm. We choose the size of 2, 4, 6 for this analysis. For \textit{few-shot classification}, as shown in Tab. \ref{numexamples}, POEM performs optimally with 4 examples. A similar trend is observed for \textit{general language understanding}, with results presented in Tab. \ref{numexamples_glu}. It is noteworthy that, in the original study \cite{touvron2023llama}, an increment in the number of examples from four to five also led to reduced accuracy. This implies that an overabundance of examples might introduce noise, thereby leading to incorrect model decisions.

%% file: paper_related.tex

\textbf{Prompt Engineering.} The traditional approach to using pre-trained LMs involves fine-tuning downstream datasets \cite{Devlin2019BERTPO, lewis2019bart}, which involves extensive updates to model parameters. However, this method has shown limited success on downstream tasks. Another approach involves utilizing manual prompts to guide LLMs in performing NLP tasks without requiring additional training \cite{brown2020language, sanh2021multitask, schick2020exploiting}. A different line of work in prompt engineering aims to develop instructional prompts, which offer task descriptions instead of fill-in-the-blank questions. In-context Learning \cite{brown2020language, liu2021makes, lu2021fantastically} achieves impressive performance by incorporating in-context demonstrations. However, these prompt engineering approaches are time-consuming and require manual tuning, which is not always feasible.

\begin{table}[t]
\centering
\begin{tabular}{{p{1cm}p{0.8cm}p{0.7cm}p{0.7cm}p{0.7cm}p{0.7cm}}}
\hline
 Parameter & Value & SST-2& CR & COLA & AG News\\
\hline
 &10  &  92.9& 90.8 & 67.2 & 79.7\\
 $N$&60  & \textbf{93.4} & \textbf{92.6} & \textbf{68.4} & \textbf{80.3}\\
 &120 & 93.3 & 92.5 & 68.2 & 80.2\\
\hline
&8 & 93.0 &90.8 & \textbf{68.7} & 80.0\\
$M$&12 & 93.1&91.0 & \textbf{68.7} & 80.2\\
&16 & \textbf{93.4}&\textbf{92.6} & 68.4 & \textbf{80.3}\\
\hline
&2 & 91.8&92.1 &67.9 & 59.4\\
$m$&4& \textbf{93.4} &\textbf{92.6}  &\textbf{68.4} & \textbf{80.3}\\
&6 &91.1 &90.8 &68.2 &71.4 \\
\hline
 &6  & 93.1 & 92.3 & 68.1&79.9\\
 $k$&10 & \textbf{93.4} & \textbf{92.6} & \textbf{68.4} & \textbf{80.3}\\
 &12 &  93.1 & 92.5 & 68.1 &79.5\\
\hline
\end{tabular}
\caption{Average accuracy across different $N$, $M$, $m$ and $k$ for few-shot text classification datasets.}
\label{tableparams2}\label{tableparams2}\label{ablationsize}\label{numexamples}\label{tableparams1}
\vspace{10pt}
\end{table}

\textbf{Prompt Optimization.} Previous studies have also explored the application of RL for prompt optimization. \cite{deng2022rlprompt} propose using RL to directly generate prompts agnostic to specific queries ; however, the generated prompts may lack meaningfulness. Another previous RL editing method \cite{zhang2022tempera} allowed editing task descriptions and in-context examples. Yet, editing descriptions seldom aids optimization, and swapping examples can be time-consuming, potentially leading back to the initial state. Moreover, the method suits categorical tasks like classification. In the realm of continuous embedding space, gradients derived from LMs are employed to directly facilitate prompt optimization, a method also referred to as \textit{soft prompt} tuning \cite{li2021prefix}. However, due to their continuous nature, \textit{soft prompts} pose challenges in comprehension \cite{lester2021power} and lack reusability across diverse models because of disparities in latent embedding spaces. With the expansion of LLMs in terms of both capacity and capabilities, there has emerged a new line of research that employs LLMs as prompt generators, prompt editors, or prompt scorers. \cite{yang2023large} propose a method utilizing LLMs as prompt optimizers, where the generated prompts rely on the prior knowledge encoded within the LLMs. \cite{zhou2022large} utilize two distinct LLMs, one as a zero-shot instruction generator and the other as a scorer to optimize instructions . However, these approaches share the drawback of relying heavily on the prior knowledge encoded within LLMs, necessitating high-quality LLMs, which can be resource-intensive. Furthermore, the variability in generated outputs by LLMs can be unpredictable and challenging to interpret.

\textbf{Examplars retrieval and ordering in In-context learning.} Research has demonstrated the significant influence of in-context example selection and arrangement on the performance of LMs. For instance, \cite{rubin2021learning} utilize a retriever to select in-context examples . Additionally, \cite{liu2021makes} propose a heuristic approach to ordering examples, ranking them based on the textual similarity between the test query and in-context examples. Recently, \cite{zhang2022tempera} have applied RL to enable the swapping of in-context examples within their action space. These studies underscore the impact of selecting and arranging in-context examples on downstream task performance. Compared to heuristics, RL solutions are theoretically guaranteed. Unfortunately, existing RL-based methods are complicated and slow during training. Our study is the first RL-based prompt optimization method that is both simple and efficient, demonstrating superior performance compared to existing counterparts.

\begin{table}[t]
\centering
\begin{tabular}{{p{1.5cm}p{0.8cm}p{0.8cm}p{1.3cm}p{1.3cm}}}
\hline
 Model & $m$ & $k$ & TruthfulQA& PIQA \\
\hline
&2 &10 &35.10 &78.67\\
&4 &10 &\textbf{38.91} & \textbf{79.27}\\
Llama 2 7B&6 &10 &38.64 &78.73 \\
\hline
&2 &10 & 36.33 &78.94\\
&4 &10 & \textbf{40.95} & \textbf{79.43}\\
Llama 2 13B&6 &10 &40.27 & 79.00\\
\hline
&4 &6  & \textbf{39.46} &  78.78\\
&4  &10 & 38.91&\textbf{79.27} \\
 Llama 2 7B &4 & 12& 38.37  &79.00\\
 \hline
&4  &6  & 40.54 &  \textbf{79.43}\\
&4  &10 & \textbf{40.95}&\textbf{79.43} \\
 Llama 2 13B &4 &12& 40.41   &78.84\\
\hline
\end{tabular}
\caption{Accuracy across different $m$ and $k$ for general language understanding datasets.}
\label{numexamples_glu}\label{tableparams1_glu}
\vspace{10pt}
\end{table}

%% file: paper_discuss.tex
We have introduced POEM, a novel approach to prompt optimization within the RL paradigm. By strategically reordering few-shot examples using episodic memory, POEM significantly enhances the performance of LLMs across a wide range of NLP tasks. Our method consistently outperforms existing techniques in few-shot classification on various datasets and demonstrates clear advancements over heuristic baselines in general language understanding tasks. The synergy between NLP and RL in POEM underscores the potential for future innovations in test-time prompt optimization algorithms, which could prove pivotal for real-world applications.
\clearpage

%% file: paper_appendix.tex
\clearpage
\section{Appendix} \label{a}

\subsection{Additional results} \label{additionalresults}

We provide additional results for comparing POEM with heuristic baselines in Tab. ~\ref{class_results_extra}. We note that our method consistently surpasses heuristic baselines. We also notice that the heuristic trend varies between different datasets.

\subsection{Training details} \label{trainingdetails}

We provide the training details in Tab. ~\ref{poemhyper}. We also include the parameters for LLMs such as top p and temperature. 

\begin{table}[h]
\centering 
\begin{tabular}{p{6cm}p{1.3cm}} 
\toprule
\multicolumn{2}{c}{Training details} \\ 
\midrule
 Minibatch size & 16 \\
 Number of iterations ($N$) & 60 \\
 Number of examples per label ($g$) & 16\\
 Number of in-context examples in total ($M$) & 16\\
 Number of training samples for non-categorical tasks & 80\\
 Number of in-context examples per prompt ($m$) & 4 \\
 Number of nearest neighbors ($k$) & 10 \\
 Positive lambda coefficient $\left(\lambda_1\right)$ & 2.0 \\
 Negative lambda coefficient $\left(\lambda_2\right)$ & 1.8 \\
 Initial epsilon value $\left(\epsilon_{initial}\right)$ & 1.0 \\
 Final epsilon value $\left(\epsilon_{final}\right)$ & 0.0001 \\
 
 LLMs' top p & 0.9\\
 LLMs' temperature & 0.6\\
\bottomrule
\end{tabular}
\caption{Hyperparameters used in our experiments.}
\label{poemhyper}
\end{table}

\subsection{Dataset details} \label{a1}

We provide information for all datasets in Tab. ~\ref{class_format} and Tab. ~\ref{class_format_glu}. For Finetuning, we use standard finetuning of the RoBERTa model from huggingface for 100 epochs, a learning rate of 0.0003 and the optimizer of Adam.

\begin{table*}[h]
\centering
\begin{tabular}{{p{2.2cm}p{2cm}p{3cm}}}
\hline
 Task & Dataset &Field name\\
\hline
& Boolq& \{question\} + \{passage\} \\
Classification & QNLI& \{text1\} + \{text2\}\\
\hline
&Hellaswag &  \{question\} \\

&PIQA & \{goal\}\\

General Language&TruthfulQA & \{question\}\\

Understanding&TriviaQA & \{question\}\\

\hline
\end{tabular}
\caption{Selected fields for nearest neighbor retrieval for datasets with multiple fields}
\label{selectedfields}
\end{table*}

\subsection{Prompting details} \label{promptingdetails}
For datasets that have more than one text field, we provide the fields used for retrieve nearest neighbor in ~\ref{selectedfields}.

\begin{table*}[htp]
    \centering 
    \begin{tabular}{p{2.5cm}p{1.1cm}p{1.1cm}p{1.1cm}p{1.2cm}p{1.1cm}p{1.1cm}p{1.1cm}p{1.1cm}} 
    \toprule
     & SST-2 & Boolq & CR & AG News & IMDB & QNLI & COLA & \textbf{Average}\\
    \midrule
    Ascending Ordering & 92.5 & 64.8 & \underline{92.4} & \textbf{80.9} & \underline{90.8} & 54.0 & 66.7 & \underline{77.4}\\
    Descending Ordering & \underline{92.7} & \underline{65.9} & 86.1 & 74.5 & 54.6 & \textbf{55.3} & \underline{67.9} & 71\\
    POEM (Ours) & \textbf{93.4(0.2)} & \textbf{66.1(0.1)} & \textbf{92.6(0.2)} & \underline{80.3(1.1)} & \textbf{90.9(0.1)} & \underline{54.3(0.2)} & \textbf{68.4(0.4)} & \textbf{78}\\
    \bottomrule
    \end{tabular}
    \caption{Accuracy and standard derivation of POEM over 4 seeds versus advanced heuristic baselines on few-shot classification. The highest accuracy is \textbf{bolded} and the second-highest accuracy is \underline{underlined}. The last column shows the average accuracy across all datasets in this table.}
    \label{class_results_extra}
    \vspace{10pt}
\end{table*}

\begin{table*}[t]
\centering
\begin{tabular}{{p{2cm}p{11cm}}}
\hline
 Dataset & Natural Instructions  \\
\hline
 SST-2& "In this task, you are given sentences from movie reviews. The task is to classify a sentence as "great" if the sentiment of the sentence is positive or as "terrible" if the sentiment of the sentence is negative. "\\
 IMDB & "In this task, you are given a review of movie. Your task is to classify given movie review into two categories: 1) great, and 2) terrible based on its content. "\\
 CR & "In this task, you are given sentences from customer reviews. The task is to classify a sentence as "great" if the sentiment of the sentence is positive or as "terrible" if the sentiment of the sentence is negative. " \\
 COLA & "You will be given a sentence. Check whether the sentence is grammatically correct and is meaningful. If the sentence is grammatically correct, then answer with "yes", otherwise answer with "no". "\\
 AG News & "Classify the news articles into the categories of World, Sports, Business, and Technology. "\\
 QNLI &  "You are given two sentences. Your task is to determine if sentence 1 entails sentence 2. If sentence 1 entails sentence 2, answer with "yes", otherwise answer with "no". "\\
 Boolq & "In this task you will be given a passage and a yes/no question based on the passage. You should answer the question using the information from the passage. "\\

\hline
\end{tabular}
\caption{Natural instructions used for few-shot classification datasets.}
\label{instructions}
\vspace{10pt}
\end{table*}

\begin{table*}[t]
\centering
\begin{tabular}{{p{2cm}p{11cm}}}
\hline
 Task & Templates\\
\hline
SST-2 & "Does the following sentence have a \{"great"\} or \{"terrible"\} sentiment? \{sentence\}. Sentiment: \{answer\_choices[label]\}"  \\

IMDB & "Review: \{sentence\}. Sentiment: \{answer\_choices[label]\}"\\

CR & "Review: \{sentence\}. Sentiment: \{answer\_choices[label]\}"\\

COLA & "I'm copy-editing a story for publication. It has the following sentence in it: \{sentence\}. Does this sentence make sense and is it grammatically correct? Please answer yes or no. Answer: \{answer\_choices[label]\}"\\

AG News & "What label best describes this news article? \{sentence\}. Answer: \{answer\_choices[label]\}"\\

QNLI & "Passage: \{passage\}. After reading this passage, I have a question: \{question\}? yes or no? Answer: \{answer\_choices[label]\}"\\

Boolq & "Passage: \{passage\}. After reading this passage, I have a question: \{question\}? True or False? \{answer\_choices[label]\}"\\

Hellaswag & "\{activity\_label\}: \{ctx\_a\} \{ctx\_b\}"\\

PIQA & "Question: \{goal\}. Answer:"\\

TruthfulQA & "Q: \{question\} A:"\\

TriviaQA & "Question: \{question\}? Answer:"\\

\hline
\end{tabular}
\caption{Templates used for in-context examples for all datasets.}
\label{templates}
\vspace{10pt}
\end{table*}

\begin{table*}[tb]
\centering 
\begin{tabular}{p{1.2cm}p{1.3cm}p{1cm}p{1cm}p{1cm}p{1.cm}p{2.4cm}p{1.9cm}} 
\toprule
\multicolumn{8}{c}{Dataset details} \\ 
\midrule
 Category & Dataset & Number of labels & |$\mathcal{D}_{train}$| & |$\mathcal{D}_{test}$| & Type & Manual Prompt & Verbalizers\\
\midrule
 & SST-2 & 2 & 32 & 1.8k& Sentiment& $\langle S\rangle$ It was $[mask]$. & great, terrible\\
 & IMDB & 2 &32 & 2k& Sentiment& It was $[mask]$.& great, terrible\\
 Single sentence& CR & 2 & 32& 2k&Sentiment & It was $[mask]$.& great, terrible\\
 & COLA & 2 &32& 1.1k& NLU&  It was $[mask]$.& yes, no \\
 & AG News & 4 & 64& 7.6k& Topic& $[mask]$ News: $\langle S\rangle$& World, Sports, Business, Tech\\
 
\midrule
 Sentence-pair& QNLI & 2 &32 &9.8k&NLI & $\langle S_1\rangle$? $[mask]$,$\langle S_2\rangle$ & yes, no\\
 & Boolq& 2 & 32 & 3.3k& MRC& $\langle S_1\rangle$? $[mask]$,$\langle S_2\rangle$ & yes, no\\

\bottomrule
\end{tabular}
\caption{Classification dataset details}
\label{class_format}
\vspace{10pt}
\end{table*}

\begin{table*}[tb]
\centering 
\begin{tabular}{p{2.0cm}p{1.3cm}p{1.3cm}p{1.3cm}p{4cm}} 
\toprule
\multicolumn{5}{c}{Dataset details} \\ 
\midrule
  Dataset & |$\mathcal{D}_{train}$| & |$\mathcal{D}_{ic}$| & |$\mathcal{D}_{test}$| & Type\\
\midrule
 Hellaswag & 16 & 16 & 10k & Commonsense Reasoning\\
 PIQA & 16 & 16 & 1.8k & Commonsense Reasoning\\
 TruthfulQA & 16 & 16 & 0.7k&Question Answering\\
 TriviaQA & 16 & 16 & 18k &Question Answering\\
 
\bottomrule
\end{tabular}
\caption{General Language Understanding dataset details}
\label{class_format_glu}
\vspace{10pt}
\end{table*}